\def\year{2022}\relax
\documentclass[letterpaper]{article} 
\usepackage{aaai22}  
\usepackage{times}  
\usepackage{helvet}  
\usepackage{courier}  
\usepackage[hyphens]{url}  
\usepackage{graphicx} 
\usepackage{natbib}  
\usepackage{caption} 
\DeclareCaptionStyle{ruled}{labelfont=normalfont,labelsep=colon,strut=off} 
\usepackage{algorithm}
\usepackage{algorithmic}

\newtheorem{MyProp}{Proposition}

\usepackage{newfloat}
\usepackage{listings}
\floatstyle{ruled}
\newfloat{listing}{tb}{lst}{}
\floatname{listing}{Listing}
\nocopyright
\usepackage{amsmath}
\usepackage{amssymb}

\usepackage{comment}
\usepackage{color}
\usepackage{bm}
\usepackage{amsfonts}
\usepackage{bbm}
\usepackage{multirow}
\usepackage{verbatim}
\usepackage{booktabs}
\usepackage{subfigure}
\usepackage{wrapfig}

\usepackage{threeparttable}

\usepackage{algorithm}
\usepackage{algorithmic}

 \usepackage{marvosym}

\usepackage{array}
\newcolumntype{H}{>{\setbox0=\hbox\bgroup}c<{\egroup}@{}}

\title{EAutoDet: Efficient Architecture Search for Object Detection}
\author {
    Xiaoxing Wang\textsuperscript{\rm 1}, 
    Jiale Lin\textsuperscript{\rm 1}, 
    Junchi Yan\textsuperscript{\rm 1}\thanks{Correspondence author is Junchi Yan. The work is in part supported by China Major State Research Development Program (2020AAA0107600), NSFC (U19B2035), and Shanghai Municipal Science and Technology Major Project (2021SHZDZX0102).}, 
    Juanping Zhao\textsuperscript{\rm 2}, 
    Xiaokang Yang\textsuperscript{\rm 1}
}
\affiliations {
    \textsuperscript{\rm 1} Department of CSE \& MoE Key Lab of Artificial Intelligence, AI Institute, Shanghai Jiao Tong University\\
     \textsuperscript{\rm 1}  Guangdong OPPO Mobile Telecommunications Co., Ltd.\\
    \{figure1\_wxx, linjiale, yanjunchi, xkyang \}@sjtu.edu.cn \quad zhaojuanping1325@oppo.com
}

\usepackage{bibentry}

\begin{document}

\maketitle

\begin{abstract}
Training CNN for detection is time-consuming due to the large dataset and complex network modules, making it hard to search architectures on detection datasets directly, which usually requires vast search costs (usually tens and even hundreds of GPU-days). In contrast, this paper introduces an efficient framework, named EAutoDet, that can discover practical backbone and FPN architectures for object detection in 1.4 GPU-days. Specifically, we construct a supernet for both backbone and FPN modules and adopt the differentiable method. To reduce the GPU memory requirement and computational cost, we propose a kernel reusing technique by sharing the weights of candidate operations on one edge and consolidating them into one convolution. A dynamic channel refinement strategy is also introduced to search channel numbers. Extensive experiments show significant efficacy and efficiency of our method. In particular, the discovered architectures surpass state-of-the-art object detection NAS methods and achieve 40.1 mAP with 120 FPS and 49.2 mAP with 41.3 FPS on COCO test-dev set. We also transfer the discovered architectures to rotation detection task, which achieve 77.05 mAP$_{\text{50}}$ on DOTA-v1.0 test set with 21.1M parameters. 
\end{abstract}

\section{Introduction}
Handcrafted neural architectures that require large amounts of trials and errors of experts to design have achieved promising performance across classification, detection, and segmentation tasks. Automated architecture search methods have been recently explored, including reinforcement learning~\cite{zoph2018learning}, evolutionary algorithm~\cite{real2019regularized}, Bayesian optimization~\cite{bananas}, as well as the more cost-effective one-shot NAS~\cite{bender2019understanding} that builds a supernet as the surrogate model to predict the performance of candidate architectures. DARTS~\cite{liu2018darts} further introduces a differentiable method that reduces the search cost to a few GPU-days. 
However, DARTS requires vast GPU memory since it has to train an over-parameterized supernet, making it impossible to search on large datasets or complex tasks.
Some works~\cite{wang2020mergenas,xu2019pc,dong2019searching} are dedicated to reducing the memory.
\begin{figure}[tb!]
    \centering
    \includegraphics[width=0.98\columnwidth]{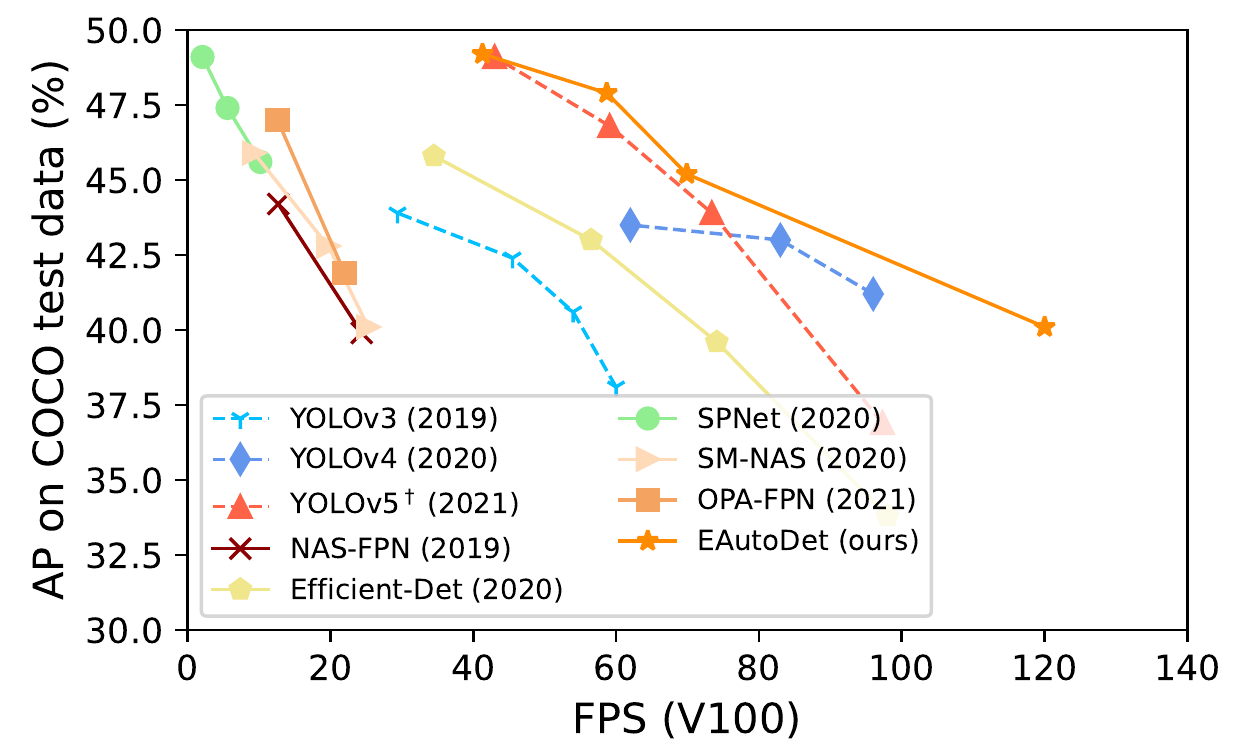}
    \vspace{-8pt}
    \caption{mAP and FPS of various detection models. Solid or dashed lines indicate NAS or handcrafted architectures. $^\dagger$: results obtained by our experiments, otherwise from the references.}
    \label{fig:result_fps}
    \vskip -0.15 in
\end{figure}

Though NAS has achieveed great success on classification tasks, it is still an open question on how to directly search architectures for detection tasks with two major difficulties: \textbf{1) It is time-consuming to train detection models from scratch on a detection dataset} due to its complex architecture compared to those for classification, which consists of multiple modules, including backbone and feature pyramid network (FPN). So that many works~\cite{girshick2015fast,bochkovskiy2020yolov4,liu2016ssd} pre-train the backbone on ImageNet; \textbf{2) Training a detection model requires vast GPU memory cost}, especially for those NAS works~\cite{chen2019detnas,hit_detector} that need to build an over-parameterized supernet. They even have to pre-train it on ImageNet, further increasing the search cost.
To simplify the supernet and lower the search difficulty, they usually restrict the search space by either searching backbone~\cite{chen2019detnas,du2020spinenet,jiang2020sp} or FPN~\cite{ghiasi2019fpn,autofpn,wang2020fcos}, which, however, actually ignores the relationship between the two modules. 
In contrast, this paper introduces kernel reusing technique and dynamic channel refinement to speed up the convergence to train a supernet and reduce GPU memory requirement. We thus propose an efficient search method, named EAutoDet, which can jointly search architectures of backbone and FPN on MS-COCO~\cite{coco} detection dataset in a few GPU-days, with no need to pre-train a supernet on ImageNet.

\begin{table*}[tb!]
    \centering
    \caption{
    Our method enables authentic fine-grained search w.r.t. operations, number of channels, and connections between layers, and is much faster than SM-NAS and Hit-Detector thanks to our kernel reusing and dynamic channel refinement techniques.
 }
    \label{tab:comparison_prior_works}
    \vspace{-8pt}
    \begin{tabular}{m{70pt}m{400pt}}
    \toprule
    \textbf{Search Space} & \textbf{Method} \\
    \midrule
    Backbone alone& DetNAS~\cite{chen2019detnas}, SpineNet~\cite{du2020spinenet}, SPNet~\cite{jiang2020sp} \\
    \hline
    FPN alone& NAS-FPN~\cite{ghiasi2019fpn}, NAS-FCOS~\cite{wang2020fcos}, OPA-FPN~\cite{liang2021opanas}, Auto-FPN~\cite{autofpn} \\
    \hline
    Joint search & SM-NAS~\cite{yao2020sm}, Hit-Detector~\cite{hit_detector}, EAutoDet (ours) \\
    \bottomrule
    \end{tabular}
\end{table*}

Additionally, prior NAS detection methods are based on RetinaNet (one-stage) or Faster-RCNN (two-stage) framework that adopts ResNet-like architectures. Few have explored to search for YOLO (one-stage) framework that could leverage its known fast speed and outstanding performance.
The handcrafted YOLO models even outperform many NAS methods in similar inference speed (shown in Table~\ref{tab:comparison_prior_works}), which implies the potential of a combination of NAS and YOLO. Nevertheless, a vanilla combination is undesirable. On the one hand, the handcrafted architecture of YOLO is subtle and elaborate. Such a nearly impeccable baseline puts forward higher requests for the ability of search method discovering the optimal architecture. On the other hand, it is better to absorb the knowledge of those well-designed architectures to design a sophisticated and large search space, which further asks for a flexible and efficient search method.
However, our method offers a practical solution thanks to its low memory requirement and rapid convergence rate to train a supernet. Besides, our technique ameliorates the computation of convolutions in the supernet rather than restricting sub-architectures of supernet, making our method flexible to suit various search spaces.
Experiments show the outstanding performance of our method on MS-COCO and DOTA-v1.0. Our contributions are summarized as follows:


\textbf{1) Efficient Architecture Search Method for Object Detection.}
We propose kernel reusing and dynamic channel refinement techniques for the fine-grained search of backbone and FPN modules in 1.4 GPU-days on a single V100 GPU, significantly outperforming prior NAS methods, e.g., 28 GPU-days~\cite{wang2020fcos} and 44 GPU-days~\cite{chen2019detnas}. Unlike other NAS methods~\cite{chen2019detnas,yao2020sm,hit_detector} that have to pre-train an over-parameterized supernet on the ImageNet, our supernet is trained from scratch on MS-COCO thanks to the property of our method: low memory requirement and rapid convergence rate to train a supernet.


\textbf{2) Sophisticated and Large Search Space for Object Detection.}
By absorbing the knowledge of well-designed YOLO models, we design a complex search space for detection, including convolution types, channel numbers, and connection of layers for backbone and FPN. It puts forward higher requests for the flexibility and efficiency of search methods. This paper offers a solution, and to the best of our knowledge, it is the first NAS method that outperforms YOLO models with competitive high inference speed.
Notice that our method can be easily applied to other detectors, e.g., Faster-RCNN~\cite{faster_rcnn} and RetinaNet~\cite{retinanet}.

\textbf{3) Strong Performance and Fast Speed.} Our design allows for direct search and evaluation without pre-training. The discovered architectures achieve outstanding performance on classic horizontal, as well as rotation detection tasks: 40.1 mAP with 120 FPS on MS-COCO where the bounding box is always assumed horizontal, and 77.05 mAP$_{50}$ on DOTA which is a dominant rotation detection benchmark but has not been used in NAS literature.

\section{Related Work}

\textbf{Object Detection.} 
Existing detection frameworks usually consist of four modules: backbone, feature fusion neck, region proposal network (in two-stage detectors), and detection head. For real-time detection, \cite{liu2016ssd,yolov1,retinanet,bochkovskiy2020yolov4,wang2021scaled} design efficient architectures for the four modules. 
There are also emerging manually-designed detectors for rotation detection whereby different loss functions are carefully devised ranging from regression~\cite{redet,yang2021rethinking} to classification~\cite{yang2021dense,csl} models. Unlike the above works requiring massive trials and expert experience to design CNNs, we aim to search architectures for detection automatically.

\textbf{Neural Architecture Search.} 
Researchers have been dedicated to efficient search algorithms for neural architectures in recent years.
NASNet~\cite{zoph2018learning} utilizes reinforcement learning (RL) and proposes to generate candidate architectures by an RNN controller. \cite{real2019regularized} adopt evolutionary algorithms (EA) to derive new architectures by crossover and mutation. Besides the above time-consuming methods, one-shot NAS~\cite{bender2019understanding} is introduced and can reduce the search cost to a few GPU-days.
DARTS~\cite{liu2018darts} regards NAS as a bi-level optimization problem and proposes to solve it by a differentiable method.
This paper adopts the differentiable method in DARTS due to its efficacy and high efficiency.

\begin{figure*}[tb!]
    \centering
    \includegraphics[width=0.8\textwidth]{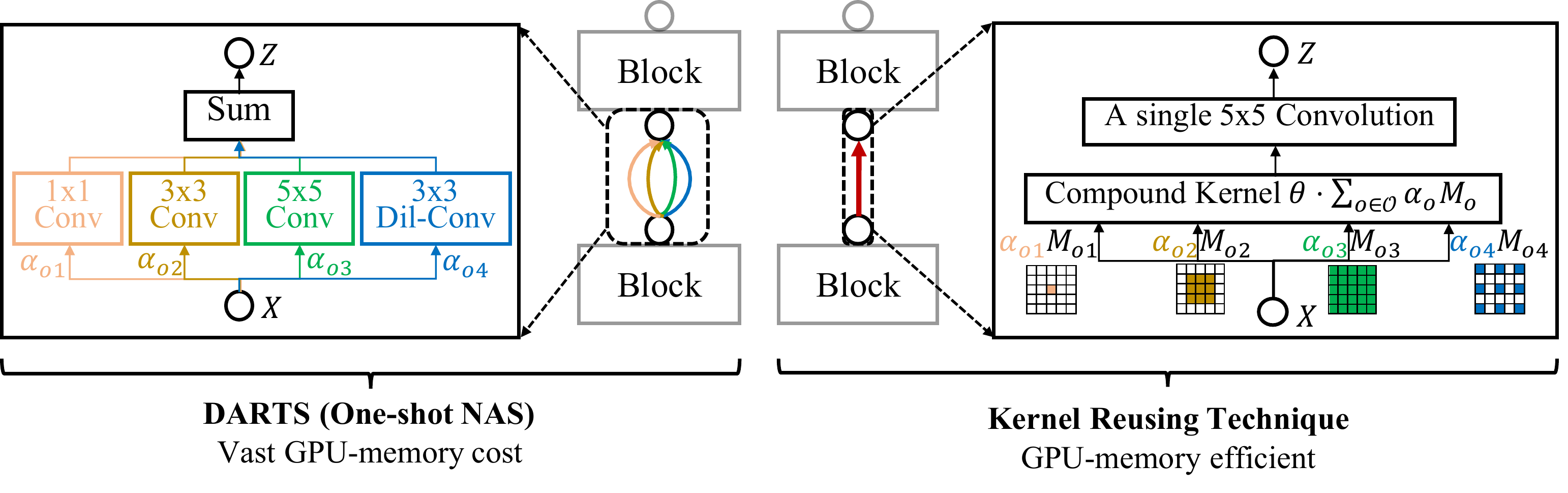}
    \vspace{-10pt}
    \caption{Compared to DARTS, the kernel reusing technique compounds multiple convolutions into a single 5x5 convolution, which can reduce the memory cost and enables efficient search for backbone and FPN. 
    }
    \label{fig:kernel_reuse}
\end{figure*}

\textbf{NAS for Object Detection.} 
Recent NAS methods for object detection can be briefly categorized into three streams: 1) Search backbone architecture and fix FPN, e.g. DetNAS~\cite{chen2019detnas} and SP-NAS~\cite{jiang2020sp}.
2) Search FPN architecture and fix backbone, e.g. NAS-FPN~\cite{ghiasi2019fpn}, Auto-FPN~\cite{autofpn}, and NAS-FCOS~\cite{wang2020fcos}. 
3) Jointly search backbone and FPN, e.g., SM-NAS~\cite{yao2020sm}, Hit-Detector~\cite{hit_detector} and our method.
Unlike SM-NAS and Hit-Detector that require vast GPUs to search, this work introduces kernel reusing and dynamic channel refinement techniques that can significantly reduce the GPU memory requirement during the search
process. Specifically, our EAutoDet can search under a more extensive search space on MS-COCO dataset directly on a single V100 GPU in 1.4 days.
Moreover, unlike many NAS detection methods~\cite{chen2019detnas,tan2020efficientdet,yao2020sm,hit_detector} that need to pre-train supernets on the ImageNet, our EAutoDet trains the supernet from scratch on the MS-COCO during the search process, demonstrating its outstanding convergence ability.


\section{The Proposed EAutoDet}
Referring to DARTS~\cite{liu2018darts} that regards NAS as a bi-level optimization task, we also build a supernet and define architecture parameters $\bm{\alpha}$ to denote the importance of candidate operations.
However, it is intractable to search on detection dataset directly by DARTS since the supernet is an over-parameterized model, making it much more challenging to train it from scratch on detection datasets, e.g., MS-COCO. We illustrate one super-edge in the supernet of DARTS in Fig.~\ref{fig:kernel_reuse} (left), which contains all independent candidate operations and thus requires massive GPU memory during the search stage.
To address the above memory explosion issue, we introduce two techniques to reduce memory requirement and computational cost: \emph{Kernel Reusing Technique} to search operation types, and \emph{Dynamic Channel Refinement} to search channel numbers.


\subsection{Kernel Reusing for Operation Type Search}
Each edge in the supernet contains multiple convolutions with various kernel sizes. Suppose $\bm{X}$ and $\bm{Z}$ are the input and output of convolutions on one edge, then $\bm{Z} = \sum_{o\in \mathcal{O}} \bm{\alpha}_o \bm{X}\otimes\bm{\theta}_o$, where `$\otimes$' denotes convolution operation, $\mathcal{O}$ is the candidate set of convolutions, and $\bm{\theta}$ is the kernel weights.

To reduce the parameters, we reuse the weights of different convolutions as shown in Fig.~\ref{fig:kernel_reuse} (right), that is, kernels of all convolutions can be extracted from unified weights by a binary mask $\bm{M}$. Moreover, since convolutions are linear operations, the weighted sum of multiple convolutions on the same edge can be compounded into one convolution. Therefore, the output of each edge can be simplified as Eq.~\ref{eq:output_edge_merge}, where $\bm{\theta}$ is the unified weights, and kernels of candidate convolution $o$ can be obtained by $\bm{M}_o\cdot\bm{\theta}$. 
\begin{equation}
    \bm{Z} = \underbrace{\sum_{o\in \mathcal{O}} \bm{\alpha}_o \bm{X}\otimes[\bm{M}_o\cdot\bm{\theta}]}_{|\mathcal{O}|\text{ convolutions}} = \underbrace{\bm{X}\otimes \left[\bm{\theta} \cdot \sum_{o\in \mathcal{O}}\bm{\alpha}_o\bm{M}_o \right]}_{\text{One convolution}}
    \label{eq:output_edge_merge}
\end{equation}

\textbf{Advantage of Our Kernel Reusing Technique.}
Apart from our kernel reusing technique, sampling-based methods~\cite{xu2019pc,dong2019searching} are also popular to reduce the memory and computational cost for classification tasks. However, they involve a dynamic network structure by sampling a sub-network of the supernet at each iteration, which will affect the supernet training. Though such an issue can be tolerated on classification tasks, it worsens on detection tasks.
In contrast, our kernel reusing technique holds a stable network structure and reduces GPU memory and computational cost by compounding multiple kernels into one without interfering with the supernet training.

\begin{figure}[tb]
    \centering
    \includegraphics[width=0.98\columnwidth]{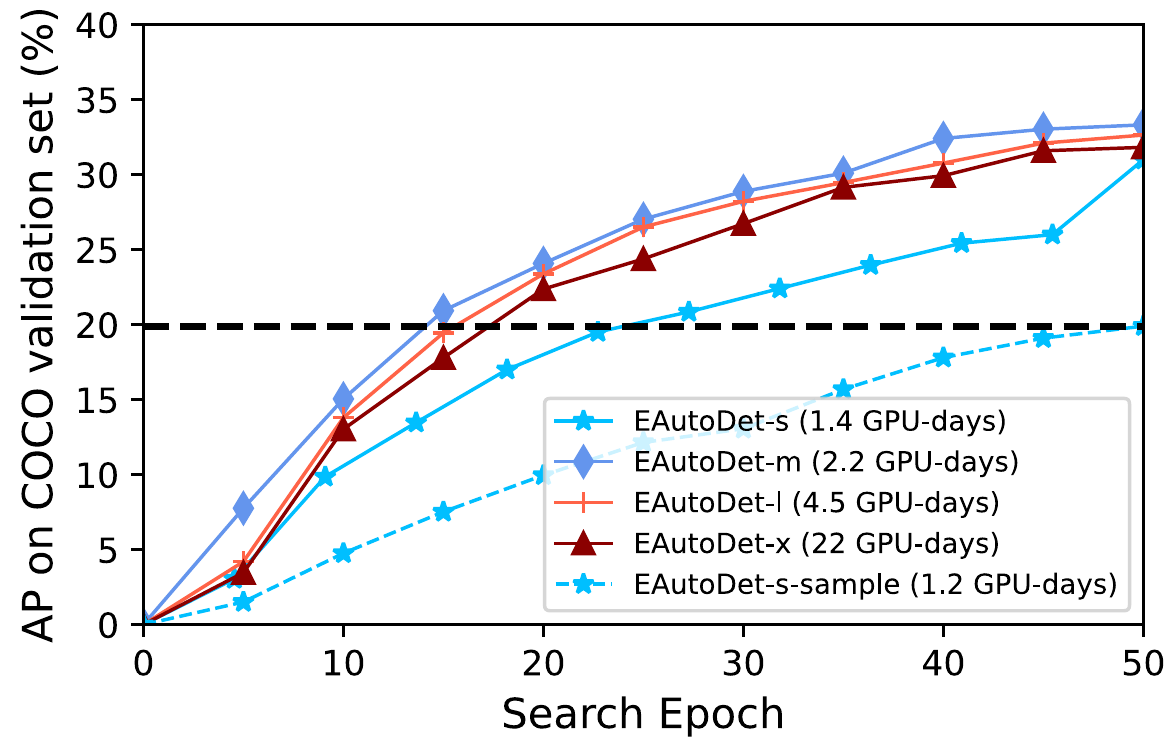}
    \vspace{-10pt}
    \caption{MAP of supernets in the search process that are trained for 50 epochs. The search cost of each supernet on a V100 GPU is given in the legend. EAutoDet-s-sample denotes search based on the sampling-based method~\cite{dong2019searching}. The horizontal dash line is the final performance of EAutoDet-s-sample.}
    \label{fig:search_tendency}
\end{figure}

We illustrate the mAP of supernets on MS-COCO validation set during the search stage in Fig.~\ref{fig:search_tendency}. EAutoDet-s/m/l/x denotes four supernets under various search spaces (details are introduced in Sec.~\ref{sec:search_space}). They are built based on our kernel reusing technique. EAutoDet-s-sample denotes that an s-level supernet is built without kernel reusing and trained by sampling operation based on Gumbel reparameterization technique at each iteration~\cite{dong2019searching}. We observe that: \textbf{1)} Our four supernets can converge to 30\% mAP; 2) Our kernel reusing technique converges better and faster than sampling-based method, which confirms the above analysis on better convergence property of our method.
Notice that Fig.~\ref{fig:search_tendency} shows mAP of supernets during 50 epochs' training in the search stage. The ultimate performance of the discovered models is evaluated by training them from scratch for 300 epochs and is reported in Table~\ref{tab:comparison_prior_works}.
The search cost is also illustrated in the legend. Specifically, an effective s-level model can be discovered in 1.4 GPU-days on a single V100 GPU, showing the remarkable efficiency of our method.

\textbf{Difference from the Prior Works.}
Unlike the prior NAS works~\cite{Stamoulis19,wang2020mergenas} that focus on classification tasks, our method aims to search detection models that are more complex and requires more computation resources. Our kernel reusing technique can significantly reduce the memory and computational cost, making it possible to discover an effective architecture in a few GPU-days.
Apart from NAS works, RepVGG~\cite{repvgg} is also related to our approach, which introduces a re-param strategy to merge skip-connection, $3\times3$ and $1\times1$ convolutions for a plain inference-time model. However, the motivation of RepVGG is to stabilize the training of VGG, and the merge process is applied after the training stage. In contrast, our kernel reusing technique is applied during the search process and aims to reduce the memory and computational cost.

\begin{figure}[tb]
    \centering
    \includegraphics[width=0.98\columnwidth]{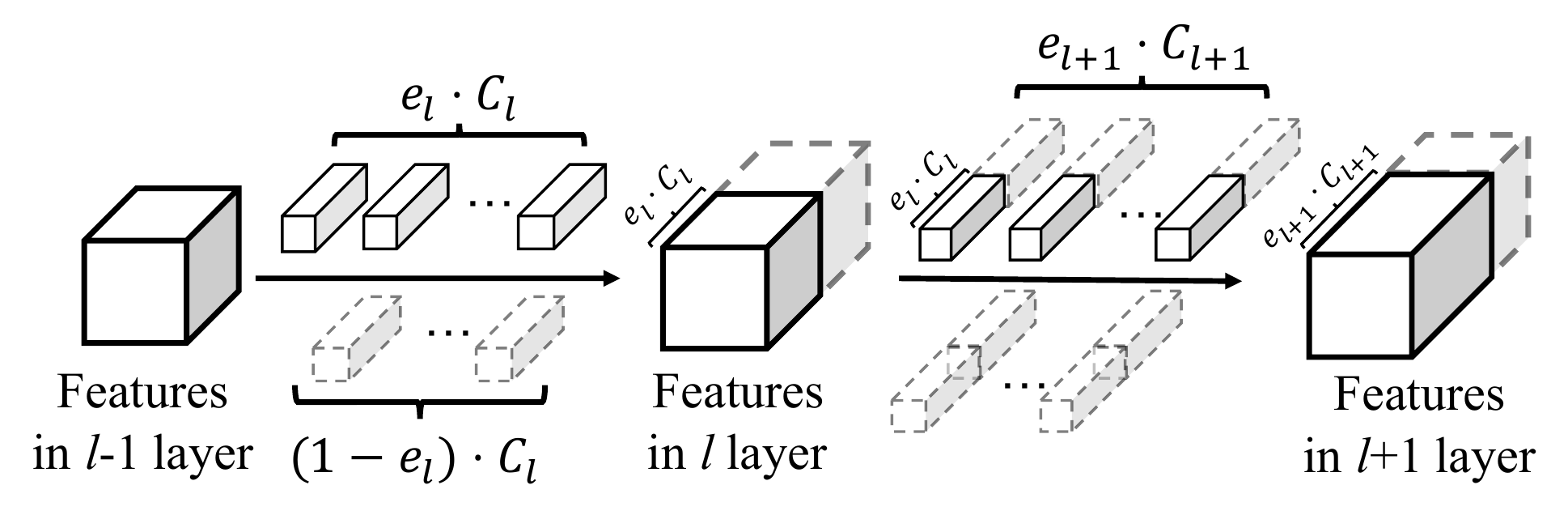}
    \vspace{-8pt}
    \caption{Refinement for convolution weights $\bm{\theta}$ on adjacent layers in our dynamic channel refinement technique. $e$ indicates the sampled expansion rate. Dotted blocks in light gray indicate the inactivated channels.}
    \label{fig:refinement_weights}
\end{figure}

\subsection{Dynamic Channel Refinement for Channel Number Search} 
Layer channels are essential hyperparameters for neural network architectures, which affect the model size and FLOPs. Unfortunately, few differentiable NAS works have explored to search channel numbers, especially for detection tasks. In this work, we introduce the dynamic channel refinement technique based on Gumbel reparameterization technique~\cite{gumbel1954statistical} to 
search for optimal expansion rates for layers. Specifically, we sample an expansion rate for each layer at every iteration and refine the operation weights $\bm{\theta}$ dynamically to fit the changeable channel numbers. 

\textbf{Sampling for Expansion Rate.}
Expansion rate for one layer can be sampled by Gumbel-argmax technique:
\begin{equation}
    \label{eq:prob_expansion}
    \bm{E} = \mathrm{one}\underline{\hbox to 0.2cm{}} \mathrm{hot} \left[\arg\max_{i}(\log {\tilde{\bm{\alpha}}_e^{i}} + \bm{g}^{i})\right],
\end{equation}
where $\tilde{\bm{\alpha}}_e = \text{softmax}(\bm{\alpha}_e)$ is the normalized weights for candidate expansion rates, and $\bm{g}^i$ are random variables sampled from $\text{Gumbel}(0,1)$ distribution. To make $\bm{E}$ differentiable w.r.t. $\bm{\alpha}_e$, we adopt Gumbel-softmax to relax the sampled vector as follows, where $\tau$ is a gradually decayed temperature.
\begin{equation}
    \label{eq:prob_expansion_softmax}
    \tilde{\bm{E}} = \frac{\exp\left[(\log\tilde{\bm{\alpha}}_e^{i}+\bm{g}^{i})/\tau\right]}   {\sum_{j} \exp\left[(\log\tilde{\bm{\alpha}}_e^{j}+\bm{g}^{j})/\tau\right]},
\end{equation}

Therefore, we adopt the one-hot vector $\bm{E}$ (Eq.~\ref{eq:prob_expansion}) to activate one candidate expansion rate during the forward pass and utilize the relaxed vector $\tilde{\bm{E}}$ ( Eq.~\ref{eq:prob_expansion_softmax}) to obtain gradients for $\bm{\alpha}_e$ during the back-propagation, which is a popular reparameterization technique that has been widely-use in many recent works ~\cite{dong2019searching,fbnetv2}.

\textbf{Dynamic Channel Refinement for Operation Weights.}
After sampling an expansion rate $e_l$ for layer $l$ with base channel number $C_l$, the output channel becomes $e_lC_l$. The operation weights on layer $l$ can be refined by preserving the first $e_lC_l$ filters. Besides, it affects the input channel of convolution on layer $l+1$. Generally, sampling channels on one layer will affect the operation weights on the current and next layer. We, therefore, dynamically refine channels for weights of adjacent layers, as shown in Fig.~\ref{fig:refinement_weights}.
FBNet-V2~\cite{fbnetv2} is related to our method, which also utilizes the Gumbel technique to search for classification model architectures. However, FBNet-V2 has to pad zero on channels to obtain a unified dimension due to short-cut connections, resulting in useless computation. In contrast, we discard the short-cut connection and dynamically refine the channel numbers for each layer at every iteration. There is no useless computation resulting from padding zeros on channels.

\textbf{Transformation for Concatenation Layers.}
The channel number for each layer alters dynamically during the search process, which brings difficulty to refine weights for convolutions after a concatenation layer. Specifically, suppose two features with expansion rates and base channels $(e_1,C_1)$ and $(e_2,C_2)$ are concatenated. A convolution is applied after the concatenation layer, then the activated input channels of weight are separated ($0\sim e_1C_1$ and $C_1\sim C_1+e_2C_2$), making it hard to extract. 
To this end, we first give the following proposition, which is proved in the supplementary material.
\begin{MyProp}
The output of a concatenation layer followed by a convolution layer is equivalent to the sum of separate convolutions on the inputs.
\label{prop:concat_sum}
\end{MyProp}
Therefore, we can transfer concatenation layers to sum layers without any loss.

\begin{figure*}[tb!]
\centering
\includegraphics[width=0.9\textwidth]{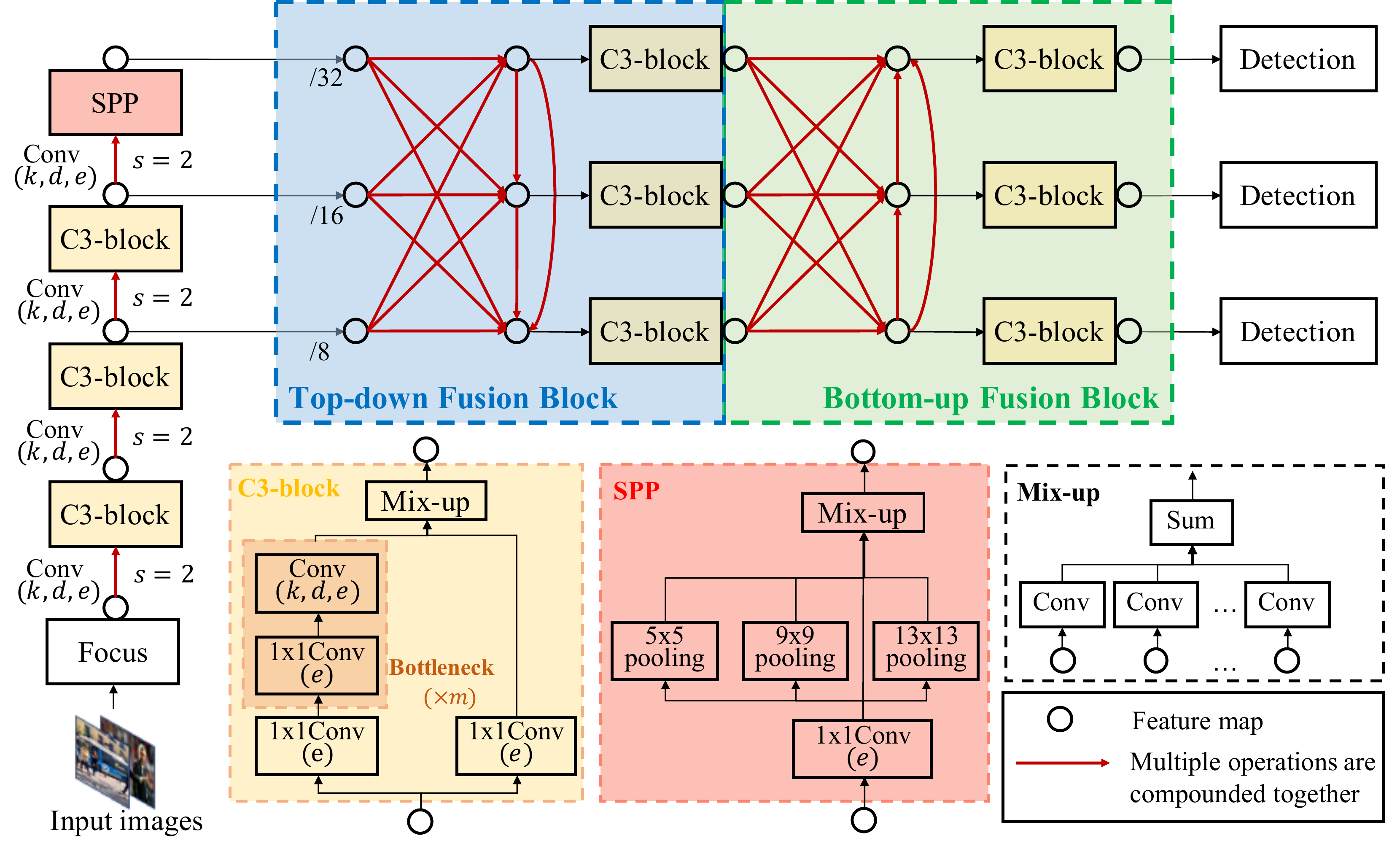}
\vspace{-15pt}
\caption{The architecture of supernet, containing all candidate operations and connections in the search space. A red edge indicates candidate operations compounded by the kernel reusing technique, which is illustrated in Fig.~\ref{fig:kernel_reuse}. In backbone, C3-block, and SPP module, parentheses under `Conv' indicate hyper-parameters to search:   kernel size $k$, dilation ratio $d$, expansion rate of output channels $e$. 
}
\label{fig:framework}
\end{figure*}

\subsection{Detection-oriented Search Space Design}
\label{sec:search_space}
\begin{table}[tb]
    \centering
    \caption{Search space size: \textbf{s}mall, \textbf{m}edium, \textbf{l}arge, e\textbf{x}tra large. Total size equals the multiplication of the backbone and FPN space sizes. Four supernets in various depth/width are designed. Details of the four search spaces are in the supplementary.}
    \label{tab:search_space_size}
    \vspace{-8pt}
    \begin{tabular}{l|ccc}
    \toprule
        \multirow{2}{*}{\textbf{Supernet}} & \multicolumn{3}{c}{\textbf{Size of Search Space}}\\
        \cmidrule{2-4}
        &\textbf{Backbone} & \textbf{FPN} & \textbf{Total} \\
    \midrule 
         EAutoDet-s & $7.9\times10^{11}$ & $9.8\times10^{24}$ & $7.7\times10^{36}$ \\
         EAutoDet-m & $3.8\times10^{18}$ & $5.2\times10^{30}$ & $2.0\times10^{49}$ \\
         EAutoDet-l & $1.8\times10^{25}$ & $2.8\times10^{36}$ & $5.0\times10^{61}$ \\
         EAutoDet-x & $8.7\times10^{31}$ & $1.5\times10^{42}$ & $1.3\times10^{74}$ \\
    \bottomrule
    \end{tabular}
\end{table}
Unlike ResNet-like and Mobile-like backbones in RetinaNet and Faster-RCNN that are transferred from classification tasks, YOLO models are specifically designed for detection tasks by experts considering both speed and performance. Therefore, we would like to absorb the knowledge of those elaborate architectures and thus design a sophisticated and large detection-oriented search space. In particular, our method ameliorates the computation of convolutions in a supernet rather than restricting architectures of sub-models, making it flexible to suit such complex and large search spaces. Specifically, we resort to YOLOv5~\cite{yolov5} and PANet~\cite{panet} and construct four types of supernets with various widths and depths, denoted as s (small), m (medium), l (large), and x (extra large), whose details are in the supplementary.
In the following, we separately introduce the search spaces for backbone and FPN for their fundamentally different roles in the detection pipeline.
The size of search space is also given in Table~\ref{tab:search_space_size}, and the details are shown in the supplementary.

\textbf{Search Space for Backbone.}
We propose to search operation types and channel numbers for down-sampling operators and structure of blocks. The supernet is shown in Fig.~\ref{fig:framework}.
\textbf{1)} Down-sampling operators have
four candidates: \{1x1 conv, 3x3 conv, 5x5 conv, 3x3 dilated conv\} and three choices of expansion rate for output channel: \{0.5, 0.75, 1.0\};
\textbf{2)} Bottleneck cell consists of two convolutions with three choices of expansion rate: \{0.5, 0.75, 1.0\}, and the second convolution have three candidates: \{3x3 conv, 5x5 conv, 3x3 dilated conv\};
\textbf{3)} C3-block has two 1x1 convolutions with two choices of expansion rates: \{0.75, 1.0\}.
Architectures of different layers are independently searched. 

\textbf{Search Space for Feature Pyramid Network.}
The supernet for top-down and bottom-up fusion blocks is shown in Fig.~\ref{fig:framework}, which enables to search connections of features in three spatial scales, operation types and channel numbers of each connection, and the structure of C3-blocks. 
\textbf{1)} Nodes indicate feature maps and connect with all their predecessors in the supernet. After the search stage, only two predecessors will be selected for each node; \textbf{2)} Each red edge contains four candidate operations: \{1x1 conv, 3x3 conv, 5x5 conv, 3x3 dilated conv\} with three possible expansion rates for the output channel: \{0.5,0.75,1.0\}; 
\textbf{3)} C3-blocks, whose architectures are also searched, are concatenated at the end of each fusion block to independently extract multi-scale features, as shown in Fig.~\ref{fig:framework}.
For each fusion block, we introduce architecture parameters $\bm{\alpha}_e$ and $\bm{\alpha}_o$ to denote the importance of edges and operations. Suppose $\tilde{\bm{\alpha}} = \text{softmax}(\bm{\alpha})$ is the normalized weight.
The fused feature $z_j = \sum_{i<j} \left[ \tilde{\alpha}^{(i,j)}_e\cdot \sum_{o\in\mathcal{O}} \tilde{\alpha}^{(i,j)}_o\cdot o(x_i) \right]$,
where $\mathcal{O}$ is the candidate operation set, $x_i$ is the features of predecessors. 
After the search stage, top-2 edges are preserved for each node according to $\bm{\alpha}_e$, and one operation is selected on each edge according to $\bm{\alpha}_o$.

\begin{figure}[tb]
    \centering
    \includegraphics[width=0.98\columnwidth]{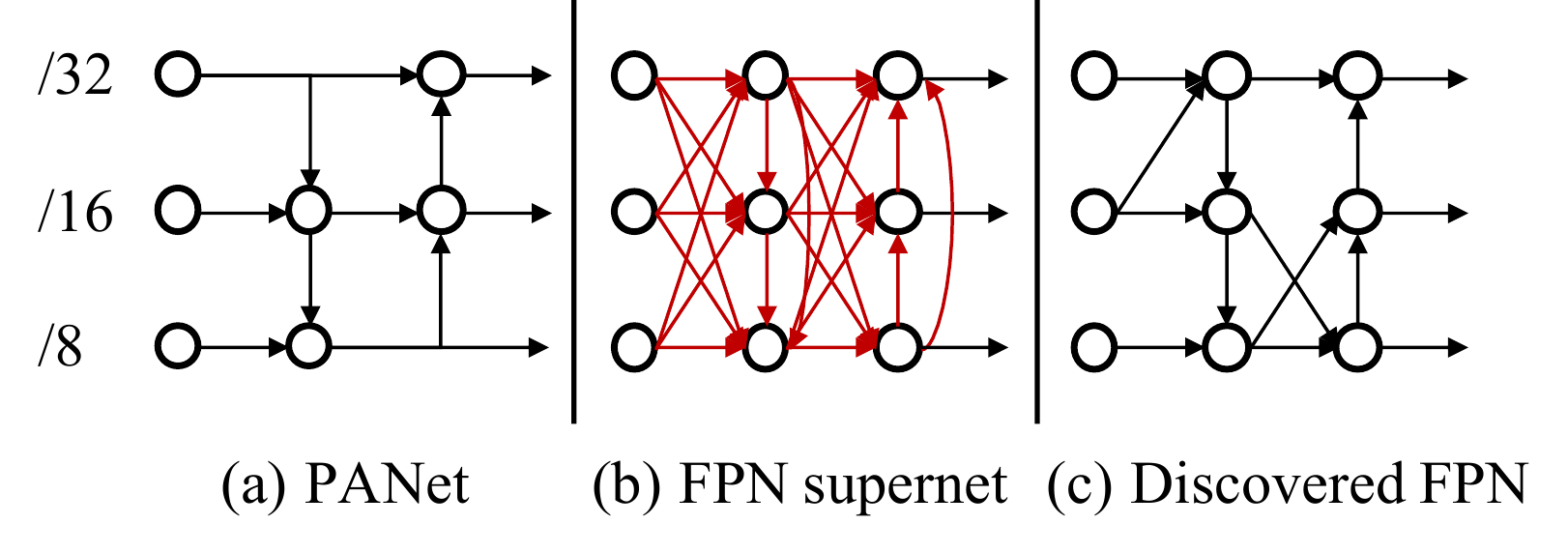}
    \vspace{-8pt}
    \caption{Architectures of PANet, supernet of FPN and our discovered model. Nodes on three rows denote feature maps on three spatial sizes. Red edges indicate multiple operations are compounded on that edge.}
    \label{fig:derive_fpn}
\end{figure}
\textbf{Deriving the Final Architecture.}
Referring to DARTS, we utilize the magnitude of architecture parameters as the importance estimation for candidate operations. The final backbone architecture is derived by preserving the best operation and removing others. While for the feature pyramid network, nodes in each fusion block preserve top-2 connections, and each connection will preserve the best operation. Fig.~\ref{fig:derive_fpn} compares the architecture of PANet, our supernet, and the discovered FPN module.

\section{Experiments}
We conduct experiments on two popular detection tasks: classic detection and rotation detection. The former is typical in many detection contests to locate common objects; The latter has been widely used in aerial images and aims to locate the ground object instances with an oriented bounding box (OBB). In this work, we first search on a large and general dataset for classic detection tasks and then evaluate the performance of our discovered architectures on both classic and rotation detection tasks.

\textbf{Benchmarks.} 
For the classic detection task, we adopt MS-COCO 2017 benchmark, which contains 118K training images, 5K validation images, and 41K test images in 80 common object categories. 
For the rotation detection task, we adopt DOTA-v1.0 benchmark, which is one of the largest aerial image detection benchmarks. The fully annotated DOTA benchmark contains 15 object categories: Plane (PL), Baseball diamond (BD), Bridge (BR), Ground field track (GTF), Small vehicle (SV), Large vehicle (LV), Ship (SH), Tennis court (TC), Basketball court (BC), Storage tank (ST), Soccer-ball field (SBF), Roundabout (RA), Harbor (HA), Swimming pool (SP), and Helicopter (HC).

\textbf{Search Settings.}
We search on MS-COCO benchmark that has sufficient common objects under various scenes.
The training set is divided into two parts to train architecture parameters and network weights.
A supernet is built in which the architectures of blocks are independently searched. The final architecture is derived after alternately optimizing architecture parameters and network weights for 50 epochs by an SGD optimizer.




\textbf{Evaluation Settings.}
We adopt mean average precision (mAP) as the evaluation metric to evaluate the performance of our models. 
Firstly, the discovered architectures are trained on MS-COCO training set from scratch for 300 epochs by an SGD optimizer and evaluated on its validation and test sets. We directly utilize the hyper-parameters provided by YOLOv5 for a fair comparison. To fairly compare the speed (FPS) with YOLO methods, we convert the trained models to the style of YOLOv4~\cite{bochkovskiy2020yolov4} and evaluate the FPS on the Darknet platform~\cite{darknet13}, which is written in C and CUDA.
Secondly, we train the architectures on rotation detection task on DOTA-v1.0 training set from scratch for 300 epochs and evaluate them on the validation and test sets. Notice that we train and test on a single input scale unlike previous works~\cite{redet,yang2021rethinking} that adopt multi-scale training technique and random rotation augmentation.
All our experiments are trained and tested on the V100 GPU, and our models are trained on PyTorch platform. 

\subsection{Ablation Study of Backbone and FPN Search}
\begin{table}[tb]
    \centering
    \caption{Performance of joint and independent search for backbone and FPN on MS-COCO validation set. }
    \label{tab:backbonevsfpn}
\vspace{-8pt}
    \begin{tabular}{llrcc}
    \toprule
       \multirow{2}{*}{\textbf{Model}} & \multicolumn{2}{c}{\textbf{Architecture}} & \multirow{2}{*}{\textbf{mAP}} & \multirow{2}{*}{$\Delta$} \\
        \cmidrule(lr){2-3}
        &\textbf{Backbone} & \textbf{FPN} & & \\
    \midrule 
         YOLOv5-s    & default   &  default    & 36.9 & +0.0 \\ 
         EAutoDet-s & \emph{searched} & default & 37.4 & +0.5 \\
         EAutoDet-s & default & \emph{searched} & 38.9 & +2.0 \\
         EAutoDet-s & \textbf{\emph{searched}}  & \textbf{\emph{searched}}  & \textbf{40.1} & \textbf{+3.2} \\
         \midrule
         YOLOv5-m    & default   &  default    & 44.0 & +0.0  \\
         EAutoDet-m & \emph{searched} & default & 44.6 & +0.6 \\
         EAutoDet-m & defualt & \emph{searched}  & 45.0 & +1.0 \\
         EAutoDet-m & \textbf{\emph{searched}}  & \textbf{\emph{searched}}  & \textbf{45.5} & \textbf{+1.5} \\

    \bottomrule
    \end{tabular}
\end{table}
We compare the performance of joint and independent search for backbone and FPN in Table~\ref{tab:backbonevsfpn}, where `default' indicates that we directly adopt the architecture of YOLOv5, and `searched' indicates that we search for the architectures.
We observe that 
1) Joint search achieves the best performance, showing the effectiveness of our algorithm and the necessity of joint search for detection models.
2) The search performance is much more sensitive to the architecture of FPN compared to the backbone module.  

\begin{table*}[tb!]
\centering
\caption{Comparison with prior works on the COCO test-dev. FPS for YOLOv5 and our method are calculated on a single V100 GPU, and results for other methods are directly obtained from their papers. Different blocks indicate models with various inference speeds and prediction performance.
`$^\dagger$': The results are obtained by our experiments. 
`-': The value is not provided by the original paper.
`$^\star$': The unit of search cost is TPU-days, while the unit of other methods is GPU-days.
`$^\ddagger$': SPNet\cite{jiang2020sp} shows the search cost on VOC is 26 GPU-days, and is six times lower than that on COCO.
}
\vskip -0.15 in
\label{tab:comparison_all}
    \resizebox{.98\linewidth}{!}{    
    \setlength{\tabcolsep}{2pt}
\begin{tabular}{llllll c c c c c c l}
\toprule
\multirow{2}{*}{\textbf{Method}} &
\multirow{2}{*}{\textbf{Backbone}} &
\multirow{2}{*}{\textbf{FPN}} &
\multirow{2}{*}{\textbf{Resolution}} &
\multirow{2}{*}{\textbf{FPS}} & \textbf{\#Params} &
\textbf{mAP} &
\textbf{AP$_{50}$} &
\textbf{AP$_{75}$} &
\textbf{AP$_S$} &
\textbf{AP$_M$} &
\textbf{AP$_L$} & \textbf{Search} \\ 
 & & & &  & \textbf{(M)} &
\textbf{(\%)} &
\textbf{(\%)} &
\textbf{(\%)} &
\textbf{(\%)} &
\textbf{(\%)} &
\textbf{(\%)} & \textbf{Cost} \\ 

\midrule
YOLOv4~\citeyearpar{bochkovskiy2020yolov4}     & CD-53  & PAN       & 416 & 96 & -   & 41.2 & 62.8 & 44.3 & 20.4 & 44.4 & 56.0 & - \\
YOLOv5s$^\dagger$~\citeyearpar{yolov5}        &  YOLOv5 & PAN   & 640 & 113 & 7.3  & 36.9 & 56.0 &    40.0  &  19.9    & 41.1      &  46.0  & -   \\
EfficientDet-D0~\citeyearpar{tan2020efficientdet} & Efficient-B0 & BiFPN   & 512 & 98 & 3.9 & 33.8 & 52.2 & 35.8 & 12.0 & 38.3 & 51.2 & - \\
NAS-FPN~\citeyearpar{ghiasi2019fpn} & Res50& Searched  &640 &24 & 60.3 &39.9 & -& -& -& -& - & 333$^\star$ \\  
NAS-FCOS@128~\citeyearpar{wang2020fcos}& Res50& Searched  & 1333$\times$800 & - & 27.8 &37.9 & -& -& -& -& - & 28 \\
SpineNet-49S~\citeyearpar{du2020spinenet}& Searched& FPN & 640 & - & 11.9 & 39.5 & 59.3 & 43.1 & 20.9 & 42.2 & 54.3 & - \\
SM-NAS:E2~\citeyearpar{yao2020sm}&\multicolumn{2}{H}{Search the combination} &800$\times$600& 25 & - & 40.0 & 58.2 & 43.4 & 21.1 & 42.4 & 51.7 & 187 \\
\textbf{EAutoDet-s (ours)}  &  \textbf{Searched}& \textbf{Searched}  &  \textbf{640}   &  \textbf{120} & \textbf{9.1}   &  \textbf{40.1}       &   \textbf{58.7}      &     \textbf{43.5 }   & \textbf{21.7}    &    \textbf{43.8}     &  \textbf{50.5}  & \textbf{1.4}    \\  
\midrule
YOLOv3 + ASFF~\citeyearpar{liu2019learning}  & D-53 &   ASFF   & 416 & 54 & -  & 40.6 & 60.6 & 45.1 & 20.3 & 44.2 & 54.1 & - \\
YOLOv4~\citeyearpar{bochkovskiy2020yolov4}     & CD-53  & PAN   & 512 & 83 & -  & 43.0 & 64.9 & 46.5 & 24.3 & 46.1 & 55.2 & - \\
YOLOv4-csp~\citeyearpar{wang2021scaled}    & CD-53& PAN  & 512 & 80$^\dagger$ & 43  & 46.2 & 64.8 & 50.2 & 24.6 & 50.4 & 61.9 &- \\
YOLOv5m$^\dagger$~\citeyearpar{yolov5}         &   YOLOv5 & PAN     & 640 & 88 & 21.4 & 43.9 & 62.5 &    47.6   &  25.1     & 48.1      &  54.9 & -    \\
EfficientDet-D1~\citeyearpar{tan2020efficientdet} & Efficient-B1 & BiFPN  & 640 & 74 & 6.6 & 39.6 & 58.6 & 42.3 & 17.9 & 44.3 & 56.0 & - \\
DetNAS~\citeyearpar{chen2019detnas}&Searched& FPN &1333$\times$800&- & - &42.0 &63.9 &45.8 &24.9 &45.1 &56.8 & 44 \\
NAS-FPN~\citeyearpar{ghiasi2019fpn} & Res50& Searched  &1024&13 & 60.3 &44.2 & -& -& -& -& - & 333$^\star$ \\
Auto-FPN~\citeyearpar{autofpn} & Res50& Searched  &800 &- & 32.6 &40.5 & 61.5 & 43.8 & 25.6 & 44.9 & 51.0 & 16 \\  
NAS-FCOS@256~\citeyearpar{wang2020fcos}&R-101& Searched &1333$\times$800&- & 57.3 &43.0 &- & -& -& -& - & 28\\
SpineNet-49~\citeyearpar{du2020spinenet}&Searched& FPN &640&- & 28.5 &42.8 &62.3 &46.1 &23.7 &45.2 &57.3 & - \\
SM-NAS:E3~\citeyearpar{yao2020sm}& \multicolumn{2}{H}{Search the combination} &800$\times$600& 20 & - &42.8 &61.2 &46.5 &23.5 &45.5 &55.6 & 187 \\
Hit-Detector~\citeyearpar{hit_detector} & Searched & Searched & 1200$\times$800 & - & 27.1 & 41.4 & 62.4 & 45.9 & 25.2 & 45.0 & 54.1 & - \\
OPA-FPN@64~\citeyearpar{liang2021opanas}&Res50& Searched &1333$\times$800&22 & 29.5 &41.9 & -& -& -& -& - & 4\\
\textbf{EAutoDet-m (ours)}  &  \textbf{Searched}& \textbf{Searched}  &   \textbf{640}  &   \textbf{70} & \textbf{28.1}   & \textbf{45.2}        &  \textbf{63.5}       & \textbf{49.1}        &    \textbf{25.7}     &  \textbf{49.1}       & \textbf{57.3}  & \textbf{2.2}      \\    
\midrule
YOLOv3 + ASFF~\citeyearpar{liu2019learning}  & D-53 & ASFF     & 608 & 46 & - & 42.4 & 63.0 & 47.4 & 25.5 & 45.7 & 52.3 & - \\
YOLOv4~\citeyearpar{bochkovskiy2020yolov4}     & CD-53 & PAN         & 608 & 62 & -   & 43.5 & 65.7 & 47.3 & 26.7 & 46.7 & 53.3 & - \\
YOLOv4-csp~\citeyearpar{wang2021scaled}    & CD-53& PAN  & 640 & 65$^\dagger$ & 53  & 47.5 & 66.2 & 51.7 & 28.2 & 51.2 & 59.8 &- \\
\textbf{EAutoDet-csp (ours)}    & \textbf{CD-53}& \textbf{Searched}  & \textbf{640} & \textbf{55}  & \textbf{49.8} & \textbf{47.8} & \textbf{66.1} & \textbf{51.9} & \textbf{28.6} & \textbf{51.5} & \textbf{60.1} & \textbf{4.2} \\
YOLOv5l$^\dagger$~\citeyearpar{yolov5}         &   YOLOv5     & PAN              & 640 & 59 & 47.1 & 46.8 & 65.4 &    50.9   &  27.7     & 51.0   &  58.5 & -    \\
EfficientDet-D2~\citeyearpar{tan2020efficientdet} & Efficient-B2 & BiFPN   & 768 & 57 & 8.1 & 43.0 & 62.3 & 46.2 & 22.5 & 47.0 & 58.4 &- \\
SPNet(BNB)~\citeyearpar{jiang2020sp}&Searched& FPN &1333$\times$800&10 & - &45.6 &64.3 &49.6 &28.4 &48.4 &60.1 & 156$^\ddagger$ \\
SM-NAS:E5~\citeyearpar{yao2020sm}& \multicolumn{2}{H}{Search the combination} &1333$\times$800& 9 & - &45.9 &64.6 &48.6 &27.1 &49.0 &58.0 & 187 \\
OPA-FPN@160~\citeyearpar{liang2021opanas}&Res50& Searched &1333$\times$800&13 & 60.6 &47.0 & -& -& -& -& - & 4 \\
\textbf{EAutoDet-l (ours)}  & \textbf{Searched}& \textbf{Searched}   &  \textbf{640}   &  \textbf{59} & \textbf{34.4}   & \textbf{47.9}        &  \textbf{66.3}       &     \textbf{52.0}    &    \textbf{28.3}     &  \textbf{52.0}       & \textbf{59.9}   & \textbf{4.5}      \\    

\midrule
YOLOv3 + ASFF~\citeyearpar{liu2019learning}  & D-53  & ASFF    & 800 & 29 & - & 43.9 & 64.1 & 49.2 & 27.0 & 46.6 & 53.4 &- \\
YOLOv5x$^\dagger$~\citeyearpar{yolov5}         &   YOLOv5         & PAN          & 640 & 43 & 87.8 & 49.1 & 67.5 &    53.6  &  30.2     & 53.4      &  61.4 &- \\
EfficientDet-D3~\citeyearpar{tan2020efficientdet} & Efficient-B3  & BiFPN  & 896 & 35 & 12 & 45.8 & 65.0 & 49.3 & 26.6 & 49.4 & 59.8 &- \\
SPNet(XB)~\citeyearpar{jiang2020sp}&Searched& FPN &1333$\times$800&6 & - &47.4 &65.7 &51.9 &29.6 &51.0 &60.4 & 156$^\ddagger$ \\
\textbf{EAutoDet-x (ours)}  & \textbf{Searched}  & \textbf{Searched} & \textbf{640}    & \textbf{41} & \textbf{86.0}    &  \textbf{49.2}      &  \textbf{67.5 }     &    \textbf{53.6}    &     \textbf{30.4}   & \textbf{53.4}       &  \textbf{61.5}   & \textbf{22}   \\   



\bottomrule
\end{tabular}
}
\end{table*}

\subsection{Results on Classic Detection Benchmark: MS-COCO}
Table~\ref{tab:comparison_all} reports the performance of our methods and compares with other state-of-the-art works on the COCO test-dev dataset.
Different blocks indicate models with various inference speeds and prediction performance. 
We observe that our discovered models (EAutoDet) achieve the best performance. Specifically, EAutoDet-s achieves $40.1$ mAP with 120 FPS, outperforming EfficientDet-D0 by 6.3\% mAP with similar inference speed.
Compared to manually-designed detectors (YOLO series), our method has competitive and even better performance.
EAutoDet-m achieves 45.2\% mAP with 69.9 FPS, surpassing YOLOv4 by 2.2\% mAP.
Moreover, we compare to YOLOv4-csp~\cite{wang2021scaled} by inheriting its backbone CD-53 and searching FPN architecture (`EAutoDet-csp' in Table~\ref{tab:comparison_all}). The discovered architecture outperforms YOLOv4-csp by 0.3\% mAP with fewer parameters.

\begin{figure*}[tb!]
\begin{center}
\subfigure[FPN of s-level]{
\includegraphics[width=0.23\textwidth]{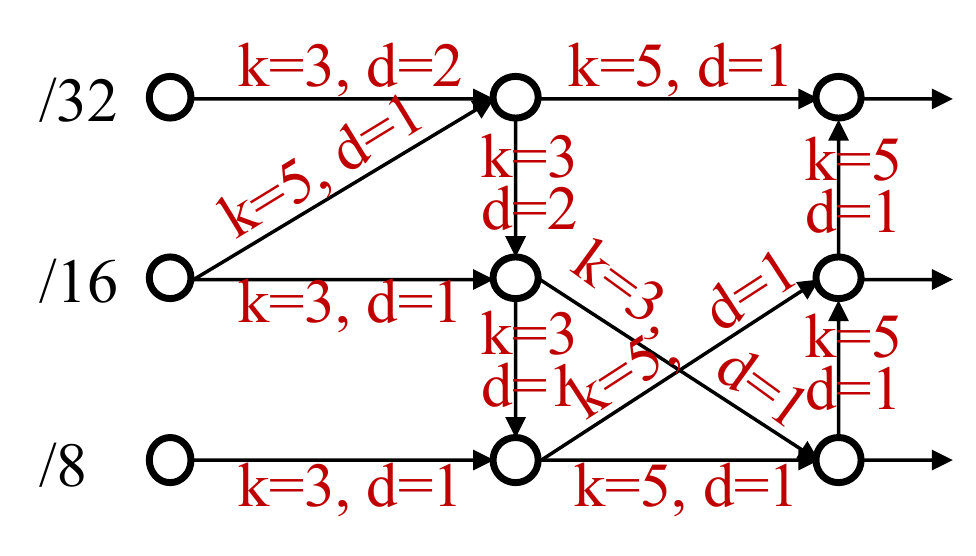}
\label{fig:tendency_num_skip}
}
\subfigure[FPN of m-level]{
\includegraphics[width=0.23\textwidth]{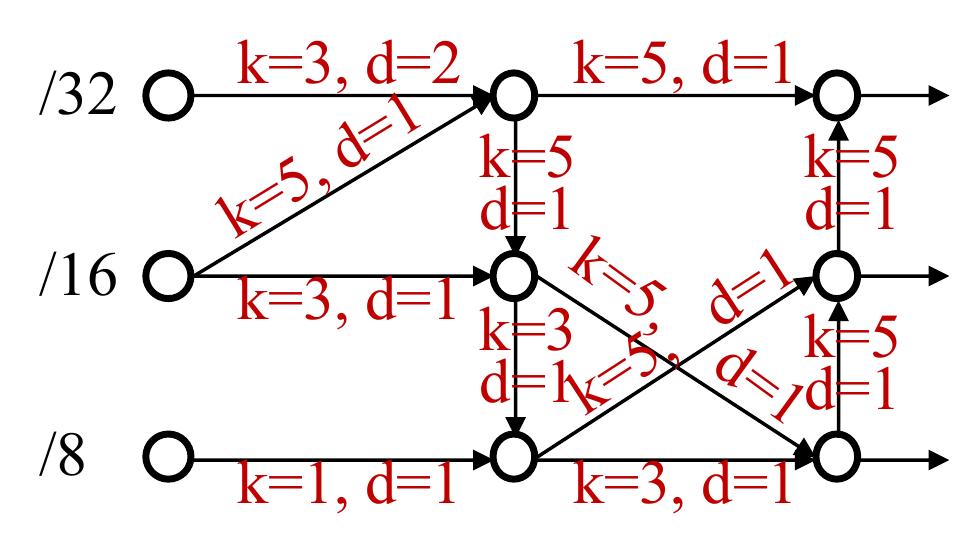}
} 
\subfigure[FPN of l-level]{
\includegraphics[width=0.23\textwidth]{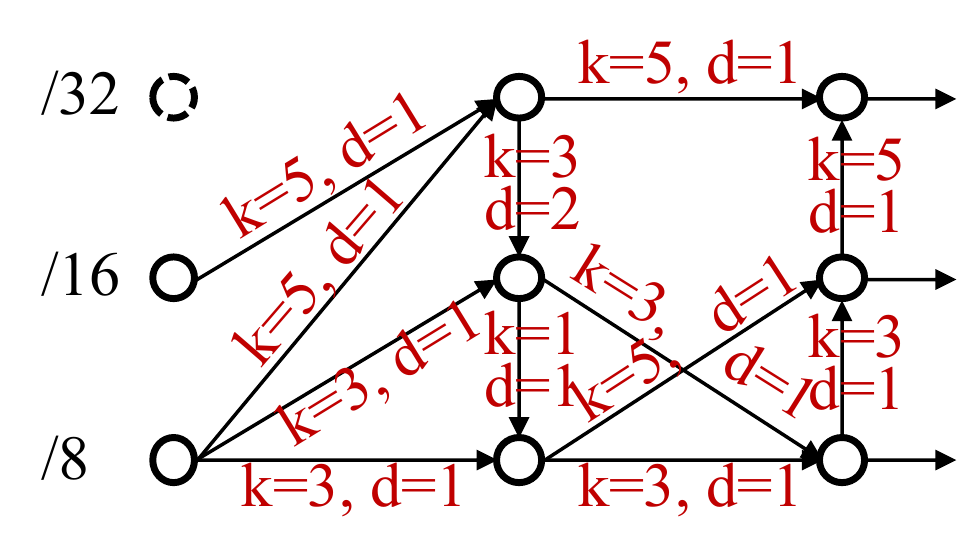}
}
\subfigure[FPN of x-level]{
\includegraphics[width=0.23\textwidth]{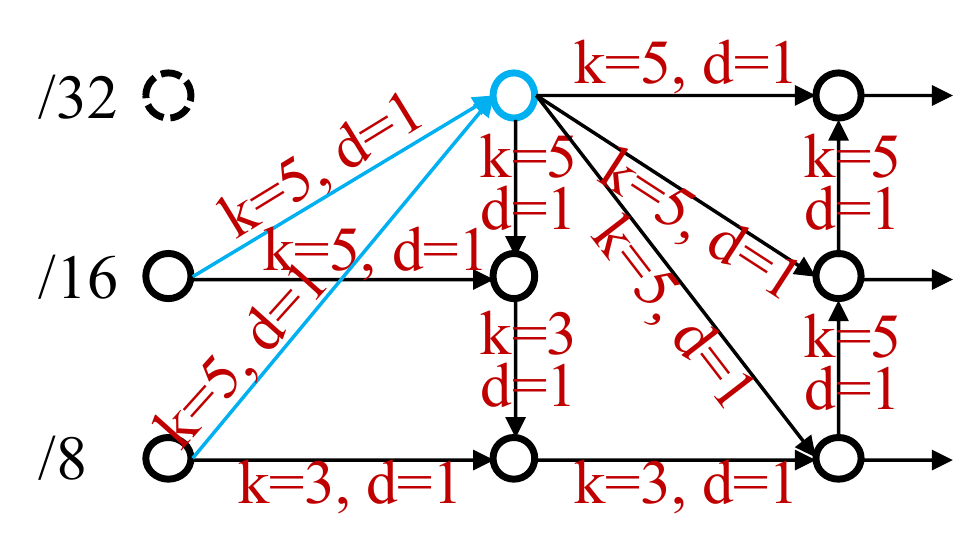}
\label{fig:fpn_architecture_x}
}
\vspace{-8pt}
\caption{FPN architecture of our searched models. `k' denotes kernel size and `d' denotes dilation ratio. Note that s-level and m-level have the same topology, while l-level and x-level discard the 32$\times$ down-sampled features (dashed node), which are the output of spatial pyramid pooling layer (SPP). In (d), we use blue lines and nodes to highlight the discarded SPP layer. We further analyze the effect of SPP by replacing one of the blue lines with the edge between dashed and blue nodes.}
\label{fig:fpn_architecture}
\end{center}
\end{figure*}

\subsection{Transferablity Evaluation}
\begin{table}[tb]
    \centering
    \caption{Comparison of the transferred models and directly searched models on the COCO validation set.}
    \label{tab:transferres}
    \vspace{-8pt}
        \begin{tabular}{lccc}
        \toprule
        \textbf{Model}    & \textbf{\#Params} & \textbf{mAP} & \textbf{$\Delta$} \\ \hline
        s (YOLOv5) & 7.3M & 36.9 & +0.0 \\
        s (transferred from x) & 8.0M & 37.4 & +0.5 \\
        s (transferred from m) & 8.0M & 39.5 & +2.6 \\
        s (searched) & 9.1M & 40.1 & +3.2 \\
        \midrule
        m (YOLOv5) & 21.4M & 44.0 & +0.0 \\
        m (transferred from x) & 21.8M & 44.6 & +0.6 \\
        m (searched) & 28.1M & 45.5 & +1.5 \\
        \midrule
        l (YOLOv5) & 47.1M & 47.0 & +0.0 \\
        l (transferred from x) & 48.6M & 47.3 & +0.3 \\
        l (searched) & 42.2M &  47.9 & +0.9 \\
        \bottomrule
        \end{tabular}
\end{table}
We transfer the discovered x-level model to s, m, and l levels to evaluate the transferability of the discovered architecture. Table~\ref{tab:transferres} compares the performance of the discovered models and the transferred models on the validation set of MS-COCO. 
We observe that: 1) Though transferred s-level and m-level can outperform baselines (YOLOv5), significant gaps exist between them and the directly searched models;  2) The transferred l-level model and the searched one achieve competitive performance. We attribute it to the different architecture preferences for small and large neural networks.
By comparing the discovered architectures, we find that they have similar backbone structures but rather different FPN structures, as shown in Fig.~\ref{fig:fpn_architecture}. Specifically, s-level and m-level have the same FPN topology but differ in operation types (kernel size and dilation ratio of convolutions). Besides, both l-level and x-level discard the 32$\times$ down-sampled features, which is the output of SPP.

To verify our analysis, we transfer the searched m-level model to s-level since they have similar FPN structures. Table~\ref{tab:transferres} shows that the s-level model transferred from m achieves 39.5\% mAP, significantly surpassing the one transferred from x by more than 2\%.


\subsection{Discussion on Spatial Pyramid Pooling}
Spatial pyramid pooling (SPP)~\cite{spp_net} is designed to integrate various receptive fields and extract multi-scale features with the same spatial size. Most recent manually-designed detectors adopt it by default, including YOLOv4 and YOLOv5. However, our experiments show that the value of SPP degrades when the network gets deeper. As shown in Fig.~\ref{fig:fpn_architecture}, models on l-level and x-level discard the SPP layer and choose to enlarge the receptive field by 5$\times$5 convolution. While models on s-level and m-level still prefer the SPP layer. 

In our analysis, SPP is vital for shallow networks as it can increase the receptive field to extract global information. However, the receptive field is enough for deep networks, making the SPP layer dispensable with the increment of network depth. Results in Table~\ref{tab:transferres} supports our analysis: When transferring x-level models to s-level, the performance degrades significantly.
To verify the effectiveness of SPP for shallow networks, we manually add SPP for the transferred s-level model. In Fig.~\ref{fig:fpn_architecture_x}, the blue node connects with 16$\times$ and $8\times$ down-sampled features. We construct two models by removing one of the connections (blue lines) and connecting the blue node with the dashed node (output of SPP layer), whose performance on the COCO validation set is reported in Table~\ref{tab:eff_spp}. We observe that after recovering the SPP layer, the performance of transferred model can be improved significantly.

\begin{table}[tb]
    \centering
    \caption{Study of SPP on MS-COCO validation set. `s-16-8' is the original transferred model from x-level, while `s-32-8' and `s-32-16' are the modified models by adding connections from the SPP layer.}
    \label{tab:eff_spp}
    \vspace{-8pt}
\begin{tabular}{lccc}
\toprule
  \textbf{Model}&\textbf{w/ SPP}&\textbf{mAP} & \textbf{$\Delta$} \\
  \midrule
  YOLOv5-s &\checkmark & 36.9 & +0.0 \\
  s-16-8 (transferred from x) & $\times$ & 37.4 & +0.5 \\
  s-32-16 (transferred from x & \checkmark & 39.0 & +2.1 \\
  s-32-8 (transferred from x) & \checkmark & 38.8 & +1.9 \\
  \bottomrule
    \end{tabular}
\end{table}

\subsection{Results on Rotation Detection Benchmark: DOTA}

\begin{table*}[tb!]
\centering
\caption{Comparison on the test set of oriented bounding boxes (OBB) task in DOTA-v1.0 benchmark. R152 denotes ResNet-152~\cite{resnet} and H-104 denotes Hourglass-104~\cite{hourglass}. $^\dagger$: Models are trained with multi-scale training techniques, while others are not. $^\star$: two-stage detection framework, while others belong to one-stage framework.
}
\vskip -0.1 in
\label{tab:obb_results}
    \resizebox{.99\textwidth}{!}{    
    \setlength{\tabcolsep}{2pt}
\begin{tabular}{lcccccccccccccccc}
\toprule
\textbf{Method} &
 \textbf{PL} &
\textbf{BD} &
\textbf{BR} &
\textbf{GTF} &
\textbf{SV} &
\textbf{LV} &
\textbf{SH} & \textbf{TC} & \textbf{BC} &\textbf{ST} & \textbf{SBF} & \textbf{RA} & \textbf{HA} & \textbf{SP} & \textbf{HC} & \textbf{mAP$_{50}$} \\
\midrule
ReDet(ReR50)~\citeyearpar{redet}$^{\star}$ & 88.79 & 82.64 & 53.97 & 74.00 & 78.13 & 84.06 & 88.04 & \textbf{90.89} & 87.78 & 85.75 & 61.76 & 60.39 & 75.96 & 68.07 & 63.39 & 76.25 \\
CenterMap(R101)~\citeyearpar{centermap}$^{\star}$ & 89.83 & 84.41 & \textbf{54.60} & 70.25 & 77.66 & 78.32 & 87.19 & 90.66 & 84.89 & 85.27 & 56.46 & 69.23 & 74.13 & 71.56 & 66.06 & 76.03 \\
CSL(R152)~\citeyearpar{csl}$^{\star}$ $^{\dagger}$ & \textbf{90.13} & 84.43 & 54.57 
& 68.13 & 77.32 & 72.98 & 85.94 & 90.74 
& 85.95 & 86.36 & 63.42 & 65.82 & 74.06 
& 73.67 & 70.08 & 76.24 \\
O$^{2}$-DNet(H104)~\citeyearpar{wei2020oriented} & 89.31 & 82.14 & 47.33 & 61.21 & 71.32 & 74.03 & 78.62 & 90.76 & 82.23 & 81.36 & 60.93 & 60.17 & 58.21 & 66.98 & 61.03 & 71.04 \\
DAL(R101)\citeyearpar{ming2021dynamic} & 88.61 & 79.69 & 46.27 & 70.37 & 65.89 & 76.10 & 78.53 & 90.84 & 79.98 & 78.41 & 58.71 & 62.02 & 69.23 & 71.32 & 60.65 & 71.78\\
P-RSDet(R101)~\citeyearpar{zhou2020arbitrary} & 88.58 & 77.83 & 50.44 & 69.29 & 71.10 & 75.79 & 78.66 & 90.88 & 80.10 & 81.71 & 57.92 & 63.03 & 66.30 & 69.77 & 63.13 & 72.30 \\
BBAVectors(R101)~\citeyearpar{yi2021oriented} & 88.35 & 79.96 & 50.69 & 62.18 & 78.43 & 78.98 & 87.94 & 90.85 & 83.58 & 84.35 & 54.13 & 60.24 & 65.22 & 64.28 & 55.70 & 72.32 \\
DRN(H104)~\citeyearpar{pan2020dynamic}$^{\dagger}$ & 89.71 & 82.34 & 47.22 & 64.10 & 76.22 & 74.43 & 85.84 & 90.57 & 86.18 & 84.89 & 57.65 & 61.93 & 69.30 & 69.63 & 58.48 & 73.23 \\
DCL(R152)~\citeyearpar{yang2021dense}$^{\dagger}$ & 89.10 & 84.13 & 50.15 & 73.57 & 71.48 & 58.13 & 78.00 & \textbf{90.89} & 86.64 & 86.78 & 67.97 & 67.25 & 65.63 & 74.06 & 67.05 & 74.06 \\
PolarDet(R101)~\citeyearpar{zhao2021polardet}$^{\dagger}$ & 89.65 & \textbf{87.07} & 48.14 & 70.97 & 78.53 & 80.34 & 87.45 & 90.76 & 85.63 & 86.87 & 61.64 & \textbf{70.32} & 71.92 & 73.09 & 67.15 & 76.64 \\
GWD(R152)~\citeyearpar{yang2021rethinking}$^{\dagger}$ &  86.96 & 83.88 & 54.36 & \textbf{77.53} & 74.41 &	68.48 &	80.34 &	86.62 &	83.41 &	85.55 &	\textbf{73.47} &	67.77 &	72.57 &	75.76 & 73.40 & 76.30 \\
\midrule
CSL(YOLOv5-s)~\citeyearpar{yolov5obb} & 88.93 & 76.07 & 50.97 & 59.61 & 81.38 & 84.00 & 88.19 & 90.81 & 80.52 & 87.63 & 47.38 & 62.85 & 68.27 & 80.63 & 62.74 & 74.00 \\
CSL(\textbf{EAutoDet-s}) & 88.64 & 84.78 & 51.02 & 62.04 & 81.30 & 84.02 & 88.23 & 90.82 & 85.79 & 87.36 & 52.20 & 65.68 & 73.72 & 74.33 & 65.93 & 75.72 \\
CSL(YOLOv5-m)~\citeyearpar{yolov5obb} & 88.36 & 85.24 & 53.86 & 62.35 & 81.19 & 84.72 & \textbf{88.57} & 90.77 & 80.98 & \textbf{88.09} & 53.92 & 61.90 & \textbf{75.99} & 80.43 & \textbf{68.59} & 76.33 \\
CSL(\textbf{EAutoDet-m}) & 89.04 & 86.61 & 53.83 & 63.05 & \textbf{81.51} & \textbf{85.12} & 88.46 & 90.77 & \textbf{88.21} & 87.93 & 56.04 & 60.96 & 75.96 & \textbf{82.13} & 66.18 & \textbf{77.05} \\
\bottomrule
\end{tabular}
}
\end{table*}

We utilize Circular Smooth Label (CSL) technique~\cite{csl} to obtain robust angular prediction through classification without suffering boundary conditions. The baseline adopts RetinaNet~\cite{retinanet} detection framework with ResNet152~\cite{resnet} backbone and FPN~\cite{lin2017feature} module. Our models are trained with rotation classification loss used in CSL~\cite{csl} from scratch for 300 epochs. Besides, we compare to YOLOv5 based on the open-sourced codes~\cite{yolov5obb}. Results are shown in Table~\ref{tab:obb_results}. We observe that: \textbf{1)} Our EAutoDet-s achieves competitive performance over many prior works in ResNet backbone;
\textbf{2)} EAutoDet-m outperforms the baseline (CSL~\cite{csl} with ResNet-152 backbone) by 0.8\% mAP$_{50}$; \textbf{3)} EAutoDet-s model surpasses original YOLOv5-s model by more than 1.7\% mAP$_{50}$ and EAutoDet-m surpasses YOLOv5-m by over 0.7\%.
Table~\ref{tab:oriented_results_yolo} further details the comparison between our models with YOLOv5.

\begin{table}[tb]
    \centering
    \caption{Comparison to YOLOv5 on the validation set of oriented bounding box (OBB) task in DOTA-v1.0 benchmark.}
    \label{tab:oriented_results_yolo}
    \vspace{-8pt}
        \begin{tabular}{lllc}
        \toprule
        \textbf{Model}    & \textbf{\#Params} & \textbf{FLOPs} & \textbf{mAP$_{50}$(\%)} \\ 
        \hline
        YOLOv5-s & 7.6M & 17.5G & 73.71 \\
        YOLOv5-m & 21.7M & 50.6G & 74.65 \\
        EAutoDet-s & 8.7M & 21.5G & 74.09 \\
        EAutoDet-m & 22.7M & 50.8G & 76.34 \\
        \bottomrule
        \end{tabular}
\end{table}
The above results show the generalization of our discovered architectures, which also verify the effectiveness of our search method. Notice that the discovered architectures are not limited to the specific CSL as tested in our experiment, as the search paradigm is agnostic to the choice of rotation detection loss, e.g., GWD~\cite{yang2021rethinking} and BBAVectors~\cite{yi2021oriented}, which we leave for future work.

\section{Conclusion}
This paper introduces kernel and dynamic channel refinement techniques and proposes a fast and memory-efficient search method for detection. We also design a sophisticated and large search space for detection by absorbing the knowledge of well-designed architectures of YOLO models. Our method can discover light-weighted models in 1.4 GPU-days, achieving 40.1 mAP on COCO test-dev with 120 FPS surpassing state-of-the-art NAS methods. Besides, our ablation studies suggest that the SPP plays a more vital role in shallow models than in deep models in the hope of facilitating future network design for detection.
Moreover, the discovered architecture archives 77.05\% mAP$_{50}$ on DOTA-v1.0 benchmark, outperforming most of the manually-designed models e.g. CSL~\cite{csl} (76.24\%), further verifying the effectiveness of our search method.

\appendix
\begin{figure}[tb]
\begin{center}
\subfigure[Input matrix]{
\includegraphics[width=0.54\columnwidth]{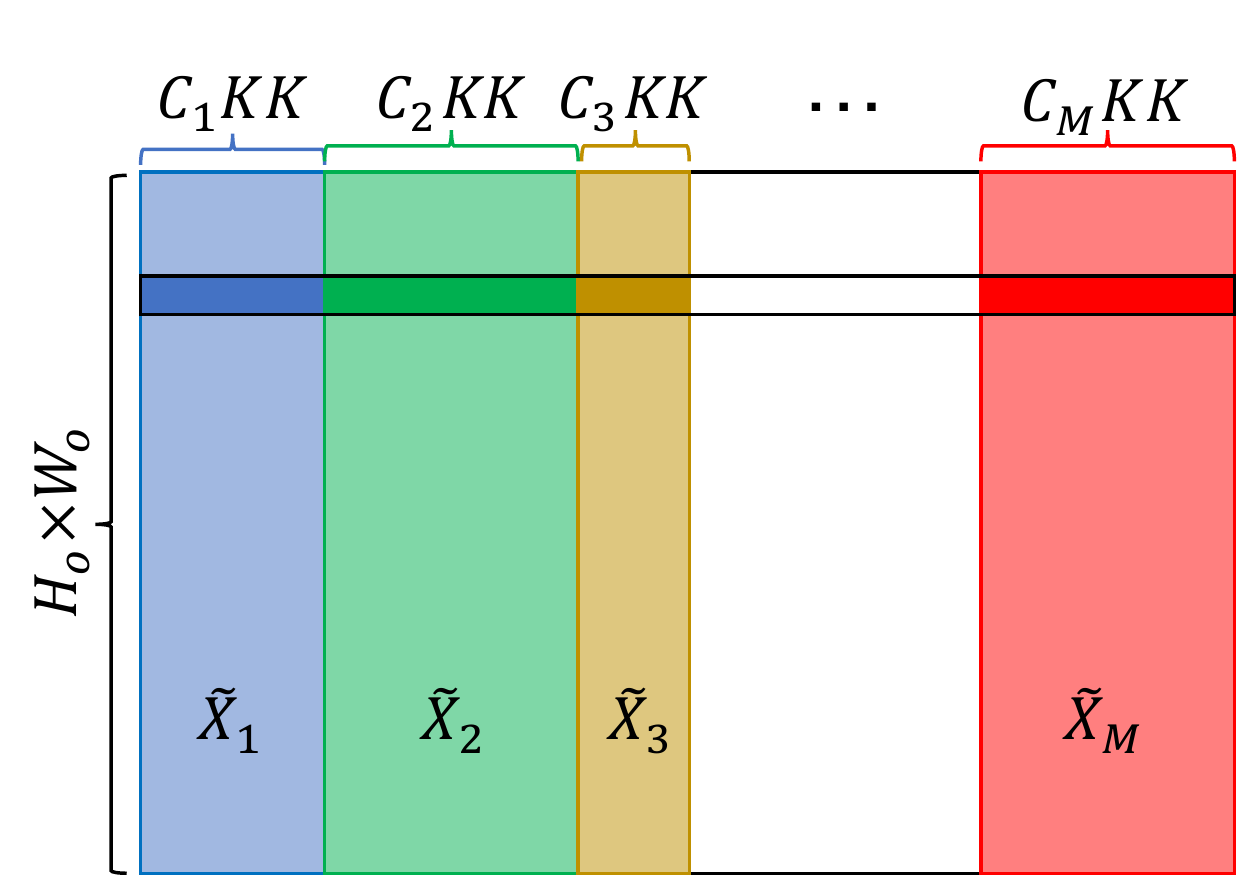}
}
\subfigure[Weight matrix]{
\includegraphics[width=0.40\columnwidth]{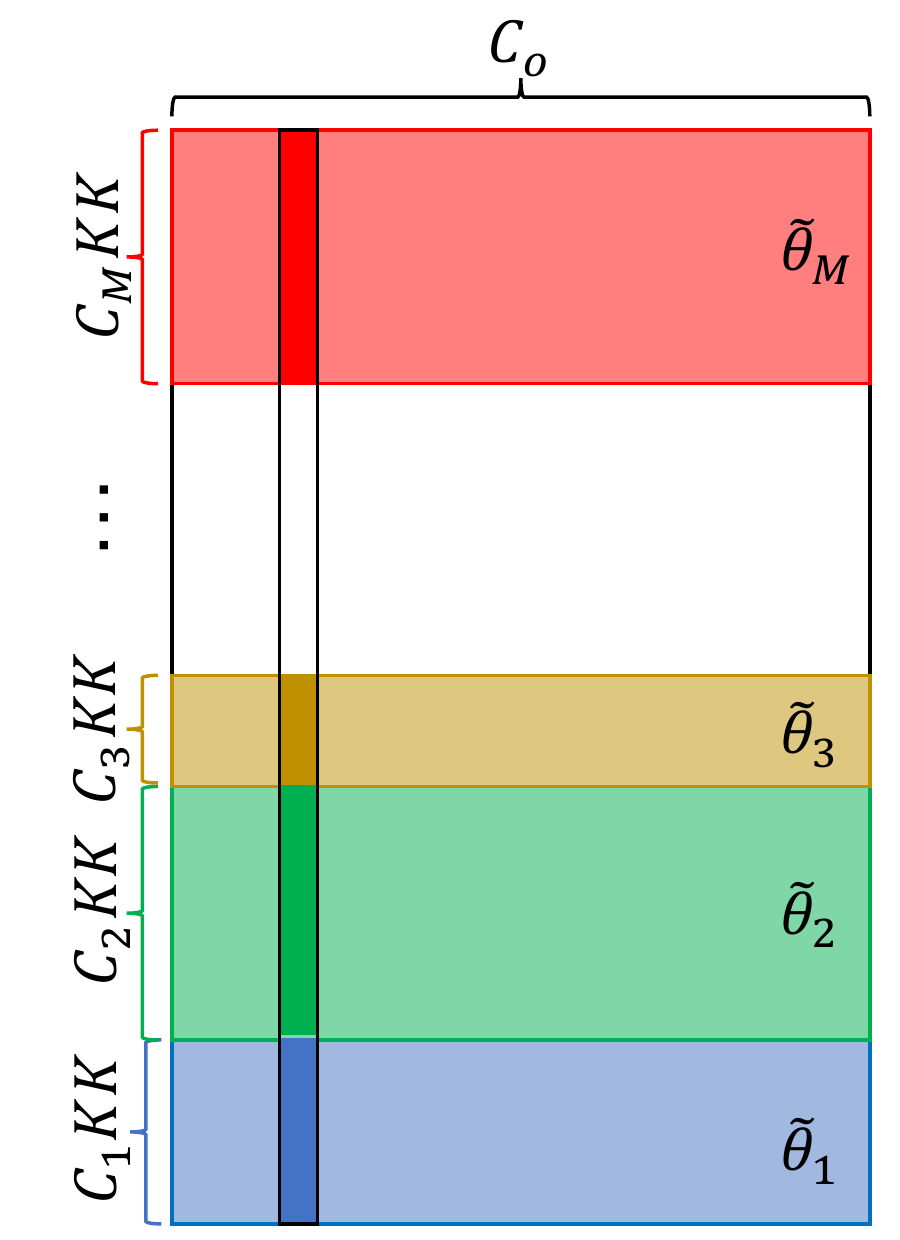}
}
\vspace{-10pt}
\caption{Illustration of an input and weight matrix $\tilde{\bm{X}}, \tilde{\bm{\theta}}$.
}
\label{fig:input_weight}
\end{center}
\end{figure} 

\section{Rationality of 
Transformation for Concatenation Layers}
To explain the rationality of transformation for concatenation layers, we can prove the following proposition.

\begin{MyProp}
The output of a concatenation layer followed by a convolution layer is equivalent to the sum of separate convolutions on the inputs.
\label{prop:concat_sum}
\end{MyProp}

Suppose we have $M$ features $\bm{X}_m\in\mathbb{R}^{N\times C_m\times H \times W}, m\in[1,M]$. By concatenating theses features along channel dimension, we can obtain a new feature $\bm{X}\in \mathbb{R}^{N\times (\sum_{m=1}^{M} C_{m})\times H\times W}$.
A convolution with weights $\bm{\theta}\in \mathbb{R}^{C_o\times (\sum_{m=1}^MC_m)\times K\times K}$ is then applied on the concatenated feature $\bm{X}$. 
Im2col operation transfers convolution to matrix multiplication, and we get the input and weight matrix $\tilde{\bm{X}}, \tilde{\bm{\theta}}$, as shown in Fig.~\ref{fig:input_weight}. $\tilde{\bm{X}} \cdot \tilde{\bm{\theta}} = \sum_{m=1}^M \tilde{\bm{X}}_m \cdot \tilde{\bm{\theta}}_m$, which is the sum of M multiplications of small matrix, where $\tilde{\bm{X}}_m \cdot \tilde{\bm{\theta}}_m$ is exactly the convolution on input feature $\bm{X}_m$. Consequently, Proposition~\ref{prop:concat_sum} is proved.

\begin{table*}[tb!]
    \caption{Structures of the four supernet built based on YOLOv5. `C' is the number of output channels (width), and `M' denotes the number of bottleneck cells in the C3 block (depth). The last line indicates the statistics of the supernet structures used to calculate the size of the search space. Specifically, $L_D, L_C, L_B$ is the number of down-sampling layers, C3-blocks and Bottlenecks in the backbone, and $K_B$ is the number of Bottlenecks in each fusion block in the FPN module.}
    \label{tab:depth_width}
\vspace{-8pt}
    \centering
    \resizebox{.98\textwidth}{!}{
    \begin{tabular}{c|l|c|c|c|c}
    \toprule
   \textbf{Module} & \textbf{Block} & \textbf{AutoYOLO-s} &\textbf{AutoYOLO-m} & \textbf{AutoYOLO-l} & \textbf{AutoYOLO-x} \\
    \midrule
    \multirow{9}{*}{Backbone} & Focus & C=32 & C=48 & C=64 & C=80 \\
    \cmidrule{2-6}
    & Down-sample & C=64 & C=96 & C=128 & C=160 \\
    \cmidrule{2-6}
    & C3 & C=64, M=1 & C=96, M=2 & C=128, M=3 & C=160, M=4 \\
     \cmidrule{2-6}
    & Down-sample & C=128 & C=192 & C=256 & C=320 \\
     \cmidrule{2-6}
    & C3 & C=128, M=3 & C=192, M=6 &  C=256, M=9 & C=320, M=12 \\
     \cmidrule{2-6}
    & Down-sample & C=256 & C=384 & C=512 & C=640 \\
     \cmidrule{2-6}
    & C3 & C=256, M=3 & C=384, M=6 & C=512, M=9 & C=640, M=12 \\
     \cmidrule{2-6}
    & Down-sample & C=512 & C=768 & C=1024 & C=1280 \\
    \cmidrule{2-6}
    & SPP & C=512 & C=768 & C=1024 & C=1280 \\
    \midrule
    \multirow{2}{*}{\shortstack[c]{Top-down \\ (FPN)}} & Feature Fusion & $\left[\begin{array}{rl}
        \text{32/} & \text{C=512} \\ \text{16/} & \text{C=256} \\ \text{8/} & \text{C=128}
    \end{array}\right]$ & $\left[\begin{array}{rl}
        \text{32/} & \text{C=768} \\ \text{16/} & \text{C=384} \\ \text{8/} & \text{C=192}
    \end{array}\right]$ & $\left[\begin{array}{rl}
        \text{32/} & \text{C=1024} \\ \text{16/} & \text{C=512} \\ \text{8/} & \text{C=256}
    \end{array}\right]$ & $\left[\begin{array}{rl}
        \text{32/} & \text{C=1280} \\ \text{16/} & \text{C=640} \\ \text{8/} & \text{C=320}
    \end{array}\right]$\\
    \cmidrule{2-6}
    & C3 & $\left[\begin{array}{rll}
        \text{32/} & \text{C=512,} & \text{M=1} \\ \text{16/} & \text{C=256,} & \text{M=1} \\ \text{8/} & \text{C=128,} & \text{M=1}
    \end{array}\right]$ & $\left[\begin{array}{rll}
        \text{32/} & \text{C=768,} & \text{M=2} \\ \text{16/} & \text{C=384,} & \text{M=2} \\ \text{8/} & \text{C=192,} & \text{M=2}
    \end{array}\right]$ & $\left[\begin{array}{rll}
        \text{32/} & \text{C=1024,} & \text{M=3} \\ \text{16/} & \text{C=512,} & \text{M=3} \\ \text{8/} & \text{C=256,} & \text{M=3}
    \end{array}\right]$ & $\left[\begin{array}{rll}
        \text{32/} & \text{C=1280,} & \text{M=4} \\ \text{16/} & \text{C=640,} & \text{M=4} \\ \text{8/} & \text{C=320,} & \text{M=4}
    \end{array}\right]$ \\
  \midrule
  \multirow{2}{*}{\shortstack[c]{Bottom-up \\ (FPN)}} & Feature Fusion & $\left[\begin{array}{rl}
        \text{32/} & \text{C=512} \\ \text{16/} & \text{C=256} \\ \text{8/} & \text{C=128}
    \end{array}\right]$ & $\left[\begin{array}{rl}
        \text{32/} & \text{C=768} \\ \text{16/} & \text{C=384} \\ \text{8/} & \text{C=192}
    \end{array}\right]$ & $\left[\begin{array}{rl}
        \text{32/} & \text{C=1024} \\ \text{16/} & \text{C=512} \\ \text{8/} & \text{C=256}
    \end{array}\right]$ & $\left[\begin{array}{rl}
        \text{32/} & \text{C=1280} \\ \text{16/} & \text{C=640} \\ \text{8/} & \text{C=320}
    \end{array}\right]$ \\
   \cmidrule{2-6}
    & C3 &$\left[\begin{array}{rll}
        \text{32/} & \text{C=512,} & \text{M=1} \\ \text{16/} & \text{C=256,} & \text{M=1} \\ \text{8/} & \text{C=128,} & \text{M=1}
    \end{array}\right]$ & $\left[\begin{array}{rll}
        \text{32/} & \text{C=768,} & \text{M=2} \\ \text{16/} & \text{C=384,} & \text{M=2} \\ \text{8/} & \text{C=192,} & \text{M=2}
    \end{array}\right]$ & $\left[\begin{array}{rll}
        \text{32/} & \text{C=1024,} & \text{M=3} \\ \text{16/} & \text{C=512,} & \text{M=3} \\ \text{8/} & \text{C=256,} & \text{M=3}
    \end{array}\right]$ & $\left[\begin{array}{rll}
        \text{32/} & \text{C=1280,} & \text{M=4} \\ \text{16/} & \text{C=640,} & \text{M=4} \\ \text{8/} & \text{C=320,} & \text{M=4}
    \end{array}\right]$  \\
    \midrule \midrule
    \multicolumn{2}{c|}{Statistics} & $L_D$=4,$L_C$=3,$L_B$=7,$K_B$=3 & $L_D$=4,$L_C$=3,$L_B$=14,$K_B$=6 & $L_D$=4,$L_C$=3,$L_B$=21,$K_B$=9 & $L_D$=4,$L_C$=3,$L_B$=28,$K_B$=12 \\
    \bottomrule
    \end{tabular}
    }
\end{table*}

\section{Details of our Search Spaces}
\textbf{Marco Architectures.}
To abosrb the knowledge of architectures of YOLO models that are well designed by experts, we refer to YOLOv5~\cite{yolov5} and build four supernets with various depths and widths, denoted as s (small), m (medium), l (large), and x (extra large). The depth and width of various types of supernet are illustrated in Table~\ref{tab:depth_width}, where `C' is the number of base output channels indicating the supernet width, and `M' is the number of bottleneck cells in the C3 block, affecting the supernet depth. The details of our search space for the backbone and FPN modules are introduced as follows.

\textbf{Backbone Search Space.}
1) For the down-sampling operator, we design
four candidates: \{1x1 convolution, 3x3 convolution, 5x5 convolution, 3x3 dilated convolution\} and three candidate expansion rates for output channel: \{0.5, 0.75, 1.0\};
2) For the bottleneck cell, which consists of two convolutions, we search output channels for both convolutions with candidates \{0.5, 0.75, 1.0\}, and kernel settings for the second convolution with candidates \{3x3 convolution, 5x5 convolution, 3x3 dilated convolution\};
3) For the C3-block, which consists of $M$ bottleneck cells, we search for the expansion rate for its output channels among two candidates: \{0.75, 1.0\}.
We adopt macro search space in this work, where architectures for bottlenecks and C3-blocks from different layers are independently searched. 
Suppose a backbone contains $L_D$ down-sampling layers, $L_C$ C3-blocks and $L_B$ Bottlenecks, the size of search space is $(4\times3)^{L_D}\cdot 2^{L_C}\cdot (3\times3)^{L_B}$.

\textbf{FPN Search Space.}
This work extracts three scales of features and builds supernets for both Top-down and Bottom-up fusion blocks. Each node in the supernet indicates a feature map connecting with all its predecessors, and each edge owns four candidate operations: \{1x1 convolution, 3x3 convolution, 5x5 convolution, 3x3 dilated convolution\} with three possible expansion rates for output channel: \{0.5,0.75,1.0\}. To derive the final architecture, each node in top-down and bottom-up fusion blocks preserves top-2 connections, and each connection will only preserve the best operation. Since we have three feature scales, there are 6 connections in each fusion block, leading to $(4\times3)^6\times\left[\binom{3}{2}\times\binom{4}{2}\times\binom{5}{2}\right]\approx5.4\times 10^8$ candidate connection types for each feature fusion block in total. 
Furthermore, we concatenate a C3 block with $K_B$ Bottlenecks after each feature fusion block, which has $2\cdot (3\times3)^{K_B}$.
Since we have two fusion blocks: Top-down and Bottom-up block, the size of search space for FPN is $\{(4\times3)^6\times\left[\binom{3}{2}\times\binom{4}{2}\times\binom{5}{2}\right]\times2\cdot (3\times3)^{K_B}\}^2$.

\section{Details of Experimental Settings}
We search on MS-COCO 2017 detection dataset~\cite{coco} and evaluation on the test set of MS-COCO and DOTA-v1.0 benchmark. All our models are trained from scratch without pre-training on ImageNet.

\textbf{Search Settings.}
We construct a supernet and define architecture parameters to represent the importance of candidate operations and connections. Unlike DARTS that shares the same cell structure, we independently search architectures for each C3 block and Bottleneck. The training set of MS-COCO is divided into two parts for training architecture parameters and network weights, respectively. The final architecture is derived after alternately optimizing architecture parameters and network weights for 50 epochs by an SGD optimizer.

\textbf{Evaluation Settings.}
The discovered architectures are trained from scratch for 300 epochs by an SGD optimizer. We directly utilize the hyper-parameters provided by YOLOv5 for a fair comparison. 
Our experiments are conducted on V100 GPU.
To fairly compare the speed (FPS) with YOLO methods, we convert the trained models to the style of YOLOv4~\cite{bochkovskiy2020yolov4} and evaluate the FPS on the Darknet platform~\cite{darknet13}, which is written in C and CUDA.
Besides, we evaluate the generalization of the discovered architectures by transferring them to rotation detection task. Specifically, we train models on  the training set of DOTA-v1.0 from scratch for 300 epochs and evaluate on the validation and test sets. Notice that we train and test on a single input scale unlike previous works~\cite{redet,yang2021rethinking} that adopt multi-scale training technique and random rotation augmentation.

\begin{table}[bt]
    \centering
    \caption{Detailed comparison between the discovered architectures and original YOLOv5 models. MAP is tested on the test set of MS-COCO.}
    \label{tab:results}
    \vspace{-8pt}
        \begin{tabular}{lcccc}
        \toprule
        \textbf{Model}    & \textbf{\#Params} & \textbf{FLOPs} & \textbf{FPS} & \textbf{mAP} \\ 
        \midrule
        YOLOv5-s & 7.3M & 17.1G & 113 & 36.9 \\
        YOLOv5-m & 21.4M & 51.4G & 88 & 43.9 \\
        YOLOv5-l & 47.1M & 112.5G & 59 & 46.8 \\
        YOLOv5-x & 87.8M & 219.0G & 43 & 49.1 \\
        \midrule
        AutoYOLO-s & 9.1M & 24.9G & 120 & 40.1 \\
        AutoYOLO-m & 28.1M & 60.8G & 70 & 45.2 \\
        AutoYOLO-l & 34.4M & 115.4G & 59 & 47.9 \\
        AutoYOLO-x & 86.0M & 225.3G & 41 & 49.2 \\
        \bottomrule
        \end{tabular}
\end{table}

\section{Comparison to YOLOv5 models.}
Table~\ref{tab:results} shows the detailed information of our models and YOLOv5 series models, including number of parameters, FLOPs, and mAP on the test test of MS-COCO. We observe that our models archieve better performance than YOLOv5 with similar parameters and FPS, showing the effectiveness of our search method. We will open-source our search and evaluation codes.

\bibliography{egbib}

\end{document}